\documentclass[letterpaper]{article} 
\usepackage{aaai25}  
\usepackage{times}  
\usepackage{helvet}  
\usepackage{courier}  
\usepackage[hyphens]{url}  
\usepackage{graphicx} 
\usepackage{pdfpages}
\usepackage{natbib}  
\usepackage{caption} 
\usepackage{algorithm}
\usepackage{algorithmic}
\usepackage{color}
\usepackage{amsmath}
\usepackage{amssymb}
\usepackage{multirow}
\usepackage{booktabs}
\usepackage{makecell}
\usepackage{rotating}
\usepackage{subfigure}

\usepackage{newfloat}
\usepackage{listings}
\DeclareCaptionStyle{ruled}{labelfont=normalfont,labelsep=colon,strut=off} 
\floatstyle{ruled}
\newfloat{listing}{tb}{lst}{}
\floatname{listing}{Listing}
\title{Forward KL Regularized Preference Optimization \\for Aligning Diffusion Policies}
\author{
    Zhao Shan\textsuperscript{\rm 1,2}\equalcontrib, Chenyou Fan\textsuperscript{\rm 1,3}\equalcontrib, Shuang Qiu\textsuperscript{\rm 4}, Jiyuan Shi\textsuperscript{\rm 1}, Chenjia Bai\textsuperscript{\rm 1, 5}\thanks{Corresponding author}
}
\affiliations{
    \textsuperscript{\rm 1}Institute of Artificial Intelligence (TeleAI), China Telecom; \textsuperscript{\rm 2}Tsinghua University;\\
    \textsuperscript{\rm 3}Northwestern Polytechnical University Xi'an; \textsuperscript{\rm 4}Hong Kong University of Science and Technology;\\
    \textsuperscript{\rm 5}ShenZhen Research Institute of Northwestern Polytechnical University\\

}

\usepackage{bibentry}

\begin{document}

\maketitle

\begin{abstract}
Diffusion models have achieved remarkable success in sequential decision-making by leveraging the highly expressive model capabilities in policy learning. A central problem for learning diffusion policies is to align the policy output with human intents in various tasks. To achieve this, previous methods conduct return-conditioned policy generation or Reinforcement Learning (RL)-based policy optimization, while they both rely on pre-defined reward functions. In this work, we propose a novel framework, Forward KL regularized Preference optimization for aligning Diffusion policies, to align the diffusion policy with preferences directly. We first train a diffusion policy from the offline dataset without considering the preference, and then align the policy to the preference data via direct preference optimization. During the alignment phase, we formulate direct preference learning in a diffusion policy, where
the forward KL regularization is employed in preference optimization to avoid generating out-of-distribution actions. We conduct extensive experiments for MetaWorld manipulation and D4RL tasks. The results show our method exhibits superior alignment with preferences and outperforms previous state-of-the-art algorithms. 
\end{abstract}

%

\section{Introduction}
In solving sequential decision-making problems, Reinforcement Learning (RL) algorithms typically adopt Gaussian or deterministic policy classes in policy optimization. Although such a policy class has successfully solved various challenging tasks \cite{bcq, BatchRL}, it can be limited in learning multi-modal policies and lead to sub-optimal behaviors in complex environments \cite{diffuser}. Recently, diffusion models \cite{ddpm, he2024learning} demonstrated superior performance compared to previous policy classes, especially for offline RL where the dataset is collected by a mixture of policies \cite{d4rl,NeoRL,yuan2024preference,fan2024task}, or the embodied manipulation tasks where the policy should imitate diverse human behaviors \cite{diffusionpolicy,humanbehavior}. Leveraging the highly expressive model capabilities, the diffusion policy exhibits strong capabilities in modeling complex behaviors. 

A central problem in learning a diffusion policy is aligning the policy output with human intents \cite{bench_off_prefer, unirlhf, yu2024regularized}. Existing methods can be roughly divided into three categories. (\romannumeral1) Several methods collect expert trajectories and directly learn a diffusion policy from the expert dataset \cite{reuss2023goal,3d-diffusion}, while they usually require a large number of expert trajectories to model human behaviors. 
To ensure the diffusion policy generates actions with desired properties, (\romannumeral2)~several methods adopt conditional generation by employing the cumulative return as the condition in modeling the trajectory \cite{yuan2024reward,decisiondiffuser}, and actions with high returns can be generated via guided sampling \cite{classifierfree,classifierguidance}. (\romannumeral3) Other works consider diffusion policy as an actor in an RL framework and perform policy improvement by maximizing cumulative rewards. Then actions can be sampled with high $Q$-values \cite{DQL,kang2024efficient}. Although the latter two methods can learn policies from an arbitrary dataset, they rely on manually defined reward functions for each task, which can be difficult to obtain in embodied tasks \cite{aligndiff,maniskill}. Motivated by recent advances in Reinforcement Learning from Human Feedback (RLHF) \cite{RLHF-1,RLHF-2} that align human preference with Large Language Models (LLMs) \cite{openAI2023GPT,touvron2023llama2}, we wonder whether the diffusion policy can be aligned with human intent directly from a collection of preference data, which removes the potentially incorrect assumption that the reward function alone drives human preferences \cite{knox2024models,rewardbench} and aims to guide the diffusion policy from preferences directly. 

In this work, we propose a novel framework, named Forward KL regularized Preference optimization for aligning Diffusion policies (\textbf{FKPD}), to align the diffusion policy with preferences. To achieve this, FKPD learns a basic diffusion policy from the offline dataset, then aligns the policy to preferences via policy optimization with a forward KL regularization with respect to the offline dataset. In the first stage, the basic policy recovers the complex action distribution in the dataset without considering the preference label, which can be multi-modal to capture the fidelity of behaviors. In the second stage, we align the basic diffusion policy to the preference data through Direct Preference Optimization (DPO) in a maximum entropy RL framework, which directly updates the policy output to match the preference model. The two stages in FKPD are complementary since the first stage learns the full distribution of actions, and the second stage searches policy based on the full distribution and converges to more narrowly distributed actions that align well with preferences. The basic policy learned in the first stage provides multi-modal policy initialization and a foundation for approximating the reverse process using the forward process for preference optimization.

The remaining challenge of this work lies in formulating the objective function for the alignment phase. To address the intractable likelihoods of diffusion policies, we approximate the optimization objective by sampling from the forward chain of the diffusion model to derive a tractable objective for preference optimization. In addition, we employ the forward KL regularization, which is more efficient for a diffusion policy than the reverse KL regularization adopted in most existing DPO literature~\cite{DPO, diffusionDPO}, to constrain the distance between the diffusion policy and the behavior policy, thereby addressing the out-of-distribution (OOD) issues in the alignment phase.

Our contribution can be summarized as follows. 
(\romannumeral1) We provide a novel framework to learn diffusion policies directly from preference data without relying on rewards and demonstrations. Specifically, we first propose a two-stage process by first learning a basic policy and then performing preference optimization on it.
(\romannumeral2) We propose the DPO objective for diffusion policies in the alignment phase which is shown to be more efficient than most existing methods~\cite{DPO, diffusionDPO}. Specifically, we employ forward KL regularization in the alignment of diffusion policies, which is shown to be more effective in addressing OOD issues.
(\romannumeral3) We conduct experiments on Meta-World \cite{metaworld} and D4RL \cite{d4rl} tasks, and the results show that FKPD exhibits superior alignment with preferences and outperforms previous state-of-the-art methods. 

\section{Preliminaries}
\subsection{Preference-based Reinforcement Learning}
We consider the general Preference-based Reinforcement Learning (PBRL) problem within a reward-free Markov Decision Process (MDP) $\mathcal{M}/r=(\mathcal{S},\mathcal{A},p,\gamma)$ with state space $\mathcal{S}$, action space $\mathcal{A}$, transition dynamics $p(s^{t+1}|s^t,a^t)$ and discount factor $\gamma$. The goal is to learn a policy $\pi(a|s)$ that maximizes an expert user's reward. We are provided with an offline dataset $\mathcal{D}$ resembling the setting of offline reinforcement learning, except that the dataset does not contain reward information. Instead, we are also given a preference dataset $\mathcal{D}_{\rm pref}=\{(\boldsymbol{\sigma}_i^+,\boldsymbol{\sigma}_i^-)\}_{i=1}^{n}$ consisting of pairs of segments sorted according to user preferences. Here $\boldsymbol{\sigma}=(s^1,a^1,s^2,a^2,...,s^k,a^k)$ is a length-$k$ segment sampled from $\mathcal{D}$. We use $\boldsymbol{\sigma}^+\succ\boldsymbol{\sigma}^-$ to indicate that segment $\boldsymbol{\sigma}^+$ is preferred to $\boldsymbol{\sigma}^-$. In the following sections, $\boldsymbol{\sigma}^+$ is referred to as the ``winning'' segment, $\boldsymbol{\sigma}^-$ is referred to as the ``losing'' segment. 

\subsection{Diffusion Model for Decision-Making}
Diffusion-based generative models \cite{ddpm} assume $p_\theta(\bold{x}_0):=\int p_\theta(\bold{x}_{0:T})d\bold{x}_{1:T}$, where $\bold{x}_1,...,\bold{x}_T$ are latent variables of the same dimensionality as the data $\bold{x}_0\sim p(\bold{x}_0)$. A forward diffusion chain gradually adds noise to the data $\bold{x}_0\sim q(\bold{x}_0)$ in $T$ steps with a pre-defined variance schedule $\beta_t$, expressed as $q(\bold{x}_{1:T}|\bold{x}_0):=\prod_{t=1}^{T}q(\bold{x}_t|\bold{x}_{t-1})$, where $q(\bold{x}_t|\bold{x}_{t-1}):=\mathcal{N}\left(\bold{x}_t;\sqrt{1-\beta_t}\bold{x}_{t-1},\beta_t\bold{I}\right)$.

A reverse chain, constructed as $p_\theta(\bold{x}_{0:T}):=\mathcal{N}(\bold{x}_T;\bold{0},\bold{I})\prod_{t=1}^{T}p_\theta(\bold{x}_{t-1}|\bold{x}_{t})$, is then optimized by maximizing the evidence lower bound (ELBO) defined as $\mathbb{E}_q\left[\ln\frac{p_\theta(\bold{x}_0)}{q(\bold{x}_{1:T}|\bold{x}_0)}\right]$. After training, sampling from the diffusion model consists of sampling from $\bold{x}_T$ and running the reverse diffusion chain from $t=T$ to $t=0$. Diffusion models can be extended to conditional generative models by conditioning $p_\theta(\bold{x}_{t-1}|\bold{x}_{t},c)$. Since there are two different types of timesteps in this work, one for the diffusion process and one for the MDP, we use superscripts to denote trajectory timesteps and subscripts to denote diffusion timesteps. 
We denote the diffusion-base policy via the reverse process of a conditional diffusion model as 
\begin{equation}
\pi_\theta(a|s)=\pi_\theta(a_{0:T}|s)=\mathcal{N}(a_T; \bold{0},\bold{I})\prod_{t=1}^{T}\pi_\theta(a_{t-1}|a_{t},s),
\end{equation}
where the end sample of the reverse chain, $a_0$, serves as the action for RL tasks. When we use a diffusion policy to fit the behavior policy on $\mathcal{D}$ (behavior clone), we adopt the simplified objective proposed by \citet{ddpm}
\begin{equation}
    \mathcal{L}_{\rm BC}=\mathbb{E}_{t\sim\mathcal{U}_{1},(s,a)\sim\mathcal{D}}\left[\left\|\boldsymbol{\epsilon}-\boldsymbol{\epsilon}_\theta(\sqrt{\bar{\alpha}_t}a+\sqrt{1-\bar{\alpha}_t}\boldsymbol{\epsilon}, s, t)\right\|^2\right],
\end{equation}
where $\boldsymbol{\epsilon}\sim\mathcal{N}(\bold{0},\bold{I})$,  $\bar{\alpha}_t$ is a scale factor defined in \citet{ddpm}.

\section{Method}
\label{sec.method}
\begin{figure*}[ht]
    \centering
    \includegraphics[width=0.7\linewidth]{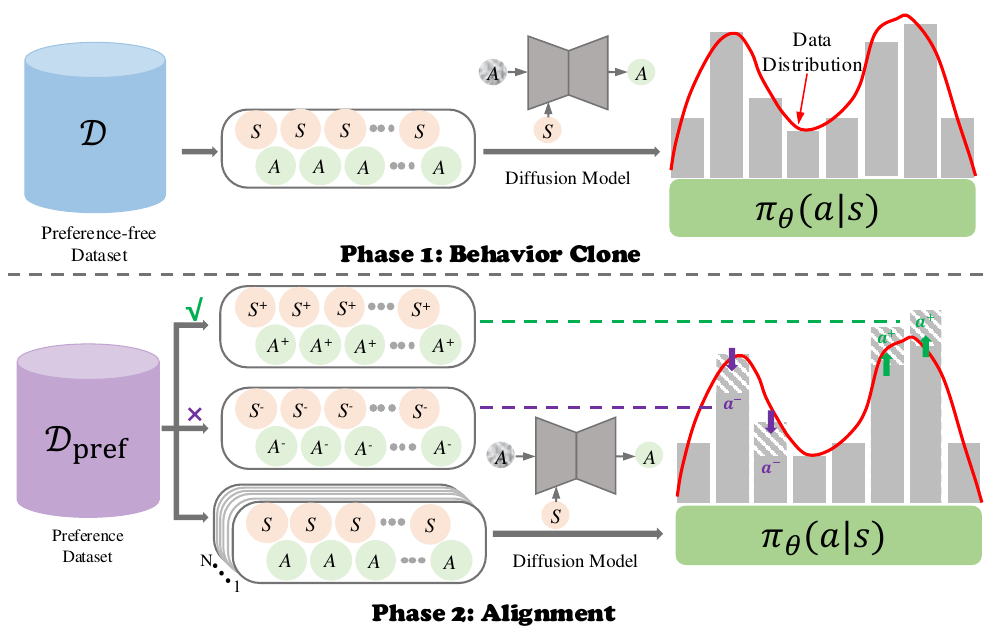}
    \caption{Overall scheme of FKPD: FKPD comprises two phases: the behavior cloning phase and the alignment phase. In the behavior clone phase, FKPD fits a diffusion policy to a preference-free dataset $\mathcal{D}$. Subsequently, in the alignment phase, FKPD employs a direct preference optimization method to align the diffusion policy with a preference dataset $\mathcal{D}_{\rm pref}$. Throughout this process, the diffusion policy is required to maintain a forward KL distance with respect to the distribution of $\mathcal{D}_{\rm pref}$}
    \label{fig.overal_scheme}
\end{figure*}

The overall scheme of our method is illustrated in Fig. \ref{fig.overal_scheme}. We first pre-train a basic diffusion policy via behavior cloning from the preference-free dataset $\mathcal{D}$, followed by aligning this policy with the preference dataset $\mathcal{D}_{\rm pref}$. For the behavior clone phase, we train a diffusion policy $\pi_\theta$ by solving $\max_{(s,a)\in\mathcal{D}}[\log \pi_\theta(a|s)]$. This is a standard process of training a diffusion policy that fits the behavior policy without reward or preference conditions~\cite{DQL} and will not be elaborated further. 
In terms of the alignment of a diffusion policy, we utilize the DPO framework, eliminating the necessity for an explicit reward model extracted from the preference dataset, thus bypassing the risk associated with an inaccurate reward model. In this section, we first give the objective of the alignment phase, referred to as $\mathcal{L}_{\rm{DPO-FK}}$, for a general generative policy from the perspective of maximum entropy RL, where we introduce a forward KL regularization to avoid the generation of OOD actions.
Subsequently, we give a practical approximation of $\mathcal{L}_{\rm {DPO-FK}}$ for a diffusion policy. Moreover, we provide an intuitive explanation of why forward KL regularization is superior to another common regularization method, namely, reverse KL regularizaion~\cite{diffusionDPO}. 
\subsection{DPO with Forward KL Regularization}
For a general generative policy, a length-$k$ segment $\sigma=(s^1,a^1,s^2,a^2,...,s^k,a^k)$ can be regarded as generated from a conditional generative model, where $(a^1,a^2,...,a^k)$ acts as the sample and $(s^1,s^2,...,s^k)$ as the condition. Here, the generative model is a stack of identical generative policies $\pi_{\theta}(a|s)$. In order to generate a segment of actions, we parallelize a stack of individual basic generative policies.
Without ambiguity, we still denote the generative model producing a segment $\boldsymbol{\sigma}$ as the generative policy $\pi_{\theta}$.  
Moreover, we omit $(s^1,s^2,...,s^k)$ in a segment and denote $\pi_{\theta}(\boldsymbol{\sigma})=\prod_{i=1}^{k}\pi_{\theta}(a^i|s^i)$ for brevity. 

During the alignment phase, we aim to optimize $\pi_{\theta}$ such that the preference score of a generated segment $\boldsymbol{\sigma}$ is maximized. 
At the same time, we also need to control $\pi_{\theta}$ to prevent it from deviating too far from the behavior policy, in order to avoid OOD issues. To this end, we use the following objective during the alignment phase:
\begin{equation}
    \label{eq.general_obj_f_k_reg}
\max_{\pi_{\theta}}\underbrace{\mathbb{E}_{\boldsymbol{\sigma}\sim\pi_{\theta}}\left[r(\boldsymbol{\sigma})\right]}_{\rm reward\:\rm maximization} -\mu\underbrace{\mathbb{D}_{\rm KL}\left[\pi_{\rm ref}\parallel\pi_{\theta}\right]}_{\rm forward\:\rm KL},
\end{equation}
where $r(\cdot)$ is a latent score function modeling human preference, 
$\pi_{\rm ref}$ is the policy obtained from the behavior cloning phase serving as a precise approximation of the behavior policy and $\mu$ is a balance factor. Here we use a forward KL regularization to control $\pi_{\theta}$ during the alignment phase. However, most existing work~\cite{diffusionDPO, DPO} employs a reverse KL regularization, where they use an objective as
\begin{equation}
    \label{eq.general_obj_r_k_reg}
\max_{\pi_{\theta}}\underbrace{\mathbb{E}_{\boldsymbol{\sigma}\sim\pi_{\theta}}\left[r(\boldsymbol{\sigma})\right]}_{\rm reward\:\rm maximization} -\mu\underbrace{\mathbb{D}_{\rm KL}\left[\pi_{\theta}\parallel\pi_{\rm ref}\right]}_{\rm reverse\: \rm KL}.
\end{equation}
Recent studies indicate that reverse KL and forward KL regularization possess distinct characteristics, with reverse KL exhibiting mode-seeking behavior and forward KL demonstrating mass-covering tendencies~\cite{sun2024inverse}. Furthermore, forward KL has been shown to better balance alignment performance and generation diversity~\cite{wang2024beyond}. In this paper, we will reveal another significant advantage of forward KL regularization in the alignment of diffusion policies: its superior effectiveness in avoiding OOD issues, which will be explained in subsequent sections.

To optimize Eq.~\eqref{eq.general_obj_f_k_reg} directly using preference data, we handle two terms in Eq.~\eqref{eq.general_obj_f_k_reg} separately. For the forward KL regularization term, we simplify it as
\begin{equation}
\begin{aligned}
    \mathbb{D}_{\rm KL}\left[\pi_{\rm ref}\parallel\pi_{\theta}\right]
    &=\mathbb{E}_{\boldsymbol{\sigma}\sim\pi_{\rm ref}}\left[\log\pi_{\rm ref}(\boldsymbol{\sigma})\right]
    -\mathbb{E}_{\boldsymbol{\sigma}\sim\pi_{\rm ref}}\left[\log\pi_{\theta}(\boldsymbol{\sigma})\right]\\
    &\approx \underbrace{\mathbb{E}_{\boldsymbol{\sigma}\sim\pi_{\rm ref}}\left[\log\pi_{\rm ref}(\boldsymbol{\sigma})\right]}_{\rm constant}
    -\mathbb{E}_{\boldsymbol{\sigma}\sim\mathcal{D}}\left[\log\pi_{\theta}(\boldsymbol{\sigma})\right].
\end{aligned}
\end{equation}
The forward KL regularization is divided into the entropy of $\pi_{\rm ref}$, which is a constant, and the cross entropy of $\pi_{\rm ref}$ and $\pi_{\theta}$. To compute the cross-entropy term, we can sample $\boldsymbol{\sigma}$ from $\mathcal{D}$ instead of from $\pi_{\rm ref}$, as $\pi_{\rm ref}$ serves as a precious approximation of behavior policy. 

In terms of the reward maximization term in~Eq.~\eqref{eq.general_obj_f_k_reg}, we consider $\pi_{\theta}$ from the perspective of maximum entropy RL~\cite{haarnoja2018soft}, leading to the following problem:
\begin{equation}
    \label{eq.max_entropy_generative_policy_objective}
\max_{\pi_{\theta}}\mathbb{E}_{\boldsymbol{\sigma}\sim\pi_{\theta}}\left[r(\boldsymbol{\sigma})-\rho \log\pi_{\theta}(\boldsymbol{\sigma})\right],
\end{equation}
where $r(\cdot)$ is a latent score function modeling human preference, and $\rho$ is a temperature parameter. This problem has a closed-form solution (see Appendix 1.1 for details), given by
\begin{equation}
\label{eq.r_pi_star}
    \pi_{\theta}^*(\boldsymbol{\sigma})=\frac{1}{Z}\text{exp} \left( \frac{r(\boldsymbol{\sigma})}{\rho} \right),
\end{equation}
where $Z$ is a normalization constant. 
In other words, the unknown score function $r$ can be expressed by the optimal generative policy as $r(\boldsymbol{\sigma})=\rho\log(\pi_\theta^*(\boldsymbol{\sigma}))+\rho\log(Z)$. According to Bradley-Terry model~\cite{BTmodel}, the human preference model on $\mathcal{D}_\text{pref}$ is written as
\begin{equation}
    \label{eq.preference_model}
    \begin{aligned}
            P(\boldsymbol{\sigma}^+\succ\boldsymbol{\sigma}^-)&=\rm{Sigmoid}(r(\boldsymbol{\sigma}^+)-r(\boldsymbol{\sigma}^-))\\&=\rm{Sigmoid}\left(\rho \log{\pi_{\theta}^*(\boldsymbol{\sigma}^+)}-\rho \log{\pi_{\theta}^*(\boldsymbol{\sigma}^-)}\right).
    \end{aligned}
\end{equation}
Then we can derive the DPO objective for a maximum entropy generative policy employing the principle of contrastive learning, i.e., minimize the cross-entropy loss of the preference model Eq.~\eqref{eq.preference_model} on $\mathcal{D}_{\text{pref}}$, which is written as
\begin{equation}
\begin{aligned}
    \label{eq.loss_dpo}
    &\mathcal{L}_{\rm{DPO}}(\pi_\theta,\mathcal{D}_{\rm{pref}})=\mathbb{E}_{(\boldsymbol{\sigma}^+,\boldsymbol{\sigma}^-)\sim\mathcal{D}_{\rm{pref}}}\\&\left[ 
-\rm{Logsigmoid}\left(
\rho \left(\log\pi_{\theta}(\boldsymbol{\sigma}^+)-\log\pi_{\theta}(\boldsymbol{\sigma}^-)
\right)\right)
\right].
\end{aligned}
\end{equation}
In the following sections, we denote $\text{Logsigmoid}(\rho(\cdot))$ as $\text{F}(\cdot)$ for the sake of brevity. Next, we combine the above derivations together and obtain the following objective for the alignment phase:
\begin{equation}
\label{eq.loss_dpo_fkl}
    \mathcal{L}_{\rm{DPO-FK}}(\pi_\theta,\mathcal{D}_{\rm {pref}})=\mathcal{L}_{\rm {DPO}}(\pi_\theta,\mathcal{D}_{\rm{pref}}) - \mu\mathbb{E}_{\boldsymbol{\sigma}\in\mathcal{D}}\log\pi_\theta(\boldsymbol{\sigma}).
\end{equation}
\subsection{Practical Approximation for Diffusion Policy}
It is fairly hard to calculate Eq.~\eqref{eq.loss_dpo_fkl} for a diffusion policy. We will next give a practical approximation. In fact, we only need to tackle the first term since the regularization term is bounded by its ELBO~\cite{ddpm}. However, the first term cannot be simplified by the same technique since the signs of the first likelihood term and the second one in Eq.~\eqref{eq.loss_dpo} are opposite. To tackle this issue, we introduce all latent variables $\boldsymbol{\sigma}_1, \boldsymbol{\sigma}_2,...,\boldsymbol{\sigma}_T$ in the reverse process similar to~\citet{diffusionDPO}. By defining the score function on the whole reverse chain as $R(\boldsymbol{\sigma}_{0:T})$, then the original score function $r$ can be calculated by marginalizing out all latent variables as 
\begin{equation}
\label{eq.r_R}
r(\boldsymbol{\sigma}_0)=\mathbb{E}_{\boldsymbol{\sigma}_{1:T}\sim\pi_\theta(\boldsymbol{\sigma}_{1:T}|\boldsymbol{\sigma}_0)}\left[R(\boldsymbol{\sigma}_{0:T})\right].
\end{equation}
Then we perform the same process as in Eqs.~\eqref{eq.max_entropy_generative_policy_objective}-\eqref{eq.loss_dpo}, yielding the DPO objective for a diffusion policy as (see Appendix 1.2 for details),
\begin{equation}
\label{eq.loss_dpo_diffusion}
\begin{aligned}
    \mathcal{L}_{\rm {DPODiff}}&(\pi_\theta,\!\mathcal{D}_{\rm{pref}})\!=\!\mathbb{E}_{\mathcal{D}_{\rm {pref}}}\Big[-\rm {F}\Big(\mathbb{E}_{\!\boldsymbol{\sigma}_{1:T}^+\sim\pi_\theta(\boldsymbol{\sigma}_{1:T}^+|\boldsymbol{\sigma}_0^+)}\\&
\log\pi_\theta(\boldsymbol{\sigma}_{0:T}^+)- \mathbb{E}_{\boldsymbol{\sigma}_{1:T}^-\sim\pi_\theta(\boldsymbol{\sigma}_{1:T}^-|\boldsymbol{\sigma}_0^-)}\log\pi_\theta(\boldsymbol{\sigma}_{0:T}^-)\Big)\Big].
\end{aligned}
\end{equation}
Now, it is also challenging to optimize Eq.~\eqref{eq.loss_dpo_diffusion} directly, as it needs to take an expectation over the whole reverse process, which is hard to track. To tackle this issue, we use the forward process $q(\boldsymbol{\sigma}_{1:T}|\boldsymbol{\sigma}_0)$ to approximate the reverse process, yielding the following objective:
\begin{equation}
\label{eq.loss_dpo_diffusion_appr}
\begin{aligned}
        &\mathcal{L}_{\rm{DPODiff}}(\pi_\theta,\mathcal{D}_{\rm {pref}})\approx \mathcal{L}_{1}(\pi_\theta,\mathcal{D}_{\rm{pref}})\\
        &\triangleq\mathbb{E}_{(\boldsymbol{\sigma}^+,\boldsymbol{\sigma}^-)\sim\mathcal{D}_{\rm{pref}}}\left[-{\rm {F}}\left(\mathbb{E}_{\boldsymbol{\sigma}_{1:T}^+\sim q(\boldsymbol{\sigma}_{1:T}^+|\boldsymbol{\sigma}_0^+)}\log\pi_\theta(\boldsymbol{\sigma}_{0:T}^+) \right.\right.\\&\left.\left.\qquad- \mathbb{E}_{\boldsymbol{\sigma}_{1:T}^-\sim q(\boldsymbol{\sigma}_{1:T}^-|\boldsymbol{\sigma}_0^-)}\log\pi_\theta(\boldsymbol{\sigma}_{0:T}^-\right)\right].
\end{aligned}
\end{equation}
This approximation makes sense for the following two considerations: on the one hand, $\pi_{\theta}$ has been trained to convergence on $\mathcal{D}$, indicating a close approximation between the backward and forward processes; on the other hand, the forward KL regularization constrains $\pi_\theta$ from deviating excessively from the pre-trained policy during the alignment phase. 

For further simplification, we decompose the reverse process in Eq.~\eqref{eq.loss_dpo_diffusion_appr} as $\pi_{\theta}(\boldsymbol{\sigma})=\pi_{\theta}(\boldsymbol{\sigma}_T)\prod_{t=1}^T\pi_{\theta}(\boldsymbol{\sigma}_{t-1}|\boldsymbol{\sigma}_{t})$, and utilize Jensen's inequality and the convexity of Logsigmoid to take the expectation over $\boldsymbol{\sigma}_t^+$ and $\boldsymbol{\sigma}_t^-$ out of Logsigmoid, and employ Gaussian reparameterization for further simplification, yielding the following bound of Eq.~\eqref{eq.loss_dpo_diffusion_appr} (Appendix 1.3), as
\begin{equation}
\label{eq.loss_dpo_diffusor_denoised}
\begin{aligned}
&\mathcal{L}_{1}(\pi_{\theta}, \mathcal{D}_{\rm {pref}})\leq\mathcal{L}_{2}(\pi_{\theta}, \mathcal{D}_{\rm {pref}})\triangleq\\&-\mathbb{E}_{(\boldsymbol{\sigma}^+,\boldsymbol{\sigma}^-)\sim\mathcal{D}_{\rm{pref}}}\mathbb{E}_{t\sim\mathcal{U}(1,T),\boldsymbol{\epsilon}_t^+,\boldsymbol{\epsilon}_t^-\sim\mathcal{N}(0,I)}\\&{\rm{F}}\left(-T\left( \left\|\boldsymbol{\epsilon}_t^+ - \boldsymbol{\epsilon}_{\theta}(\boldsymbol{\sigma}_t^+,t)\right\|^2 - \left\|\boldsymbol{\epsilon}_t^- - \boldsymbol{\epsilon}_{\theta}(\boldsymbol{\sigma}_t^-,t)\right\|^2 \right)\right),
\end{aligned}
\end{equation}
where $\boldsymbol{\sigma}_t=\sqrt{\bar{\alpha}_t}\boldsymbol{\sigma}_0+\sqrt{1-\bar{\alpha}_t}\boldsymbol{\epsilon}$. It is obvious that the constant $T$ can be factored into $\rho$. 
We next substitute Eq.~\eqref{eq.loss_dpo_diffusor_denoised} into Eq.~\eqref{eq.loss_dpo_fkl} and bound the regularization term in Eq.~\eqref{eq.loss_dpo_fkl} using its ELBO, resulting in the complete objective:
\begin{equation}
    \label{eq.loss_align_1}
    \begin{aligned}
        &\mathcal{L}_{\rm DPODiffFK}(\pi_\theta,\!\mathcal{D}_{\rm {pref}})\!=\!\mathcal{L}_{2}(\pi_{\theta}, \!\mathcal{D}_{\rm{pref}})\\ &+\mu\mathbb{E}_{\boldsymbol{\sigma}\in\mathcal{D}}\mathbb{E}_{t\sim\mathcal{U}(1,T),\boldsymbol{\epsilon}_t\sim\mathcal{N}(0,I)}\left(\left\|\boldsymbol{\epsilon}_t - \boldsymbol{\epsilon}_{\theta}(\boldsymbol{\sigma}_t,t)\right\|^2\right).
    \end{aligned}
\end{equation}
Finally, we move the second term of Eq.~\eqref{eq.loss_align_1} into Logsigmoid, leading to our final loss for the alignment phase:
\begin{equation}
\label{eq.loss_alignment}
    \begin{aligned}
    &\mathcal{L}_{\rm{FKPD}}(\pi_{\theta}, \mathcal{D}_{\rm {pref}})=-\mathbb{E}_{(\boldsymbol{\sigma}^+,\boldsymbol{\sigma}^-)\sim\mathcal{D}_{\rm{pref}}}\mathbb{E}_{t\sim\mathcal{U}(1,T),\boldsymbol{\epsilon}_t^+,\boldsymbol{\epsilon}_t^-\sim\mathcal{N}(0,I)}\\&\rm{Sigmoid}\left(-\rho\left(\underbrace{\left\|\boldsymbol{\epsilon}_t^+ - \boldsymbol{\epsilon}_{\theta}(\boldsymbol{\sigma}_t^+,t)\right\|^2 - \left\|\boldsymbol{\epsilon}_t^- - \boldsymbol{\epsilon}_{\theta}(\boldsymbol{\sigma}_t^-,t)\right\|^2}_{\text{preference}} \right.\right.\\&\quad\quad\quad\left.\left.+ \mu\underbrace{\mathbb{E}_{\boldsymbol{\sigma}\in\mathcal{D}}\mathbb{E}_{\boldsymbol{\epsilon}_t\sim\mathcal{N}(0,I)}\left\|\boldsymbol{\epsilon}_t - \boldsymbol{\epsilon}_{\theta}(\boldsymbol{\sigma}_t,t)\right\|^2-b}_{\text{regularization}} \right)\right),
\end{aligned}
\end{equation}
where $b$ is a constant bias term to ensure the Sigmoid function operates within its linear regime. It should be noted that we perform the final merge solely to facilitate the finding of the hyperparameter $\mu$ since the preference term and the regularization term operate at the same scale. In practice, the regularization term can be constrained in a small regime around zero, which will not compromise the accuracy of the Bradley-Terry preference model.

Now we can provide an intuitive explanation of how $\mathcal{L}_{\text{FKDP}}$ aligns the pretrained behavior clone policy with $\mathcal{D}_{\text{pref}}$. We first define $\mathbb{E}_{\boldsymbol{\sigma},t}\left\|\boldsymbol{\epsilon}_t - \boldsymbol{\epsilon}_{\theta}(\boldsymbol{\sigma}_t,t)\right\|^2$ as the Denoising Mean Square Error, abbreviated as \textbf{D-MSE}. A smaller D-MSE for a segment implies a higher likelihood that the diffusion policy will generate that segment. There are two terms inside the Logsigmoid function: the preference term describes the discrepancy in the D-MSE for the winning segment relative to the losing segment, and the regularization term represents the average D-MSE for dataset $\mathcal{D}$. Since the Sigmoid function is monotonic, minimizing Eq.~\eqref{eq.loss_alignment} implies minimizing the sum of preference term and regularization term. 
As a result, objective Eq.~\eqref{eq.loss_alignment} guides $\pi_{\theta}$ in a direction that 
increases the difference between the D-MSE of losing segments and that of the winning segments, while maintaining the overall D-MSE for both winning and losing segments. This is equivalent to increasing the likelihood of winning segments relative to the likelihood of losing segments while holding the likelihood of the whole dataset, which is exactly the goal of alignment phase.

\subsection{Advantage over Reverse KL Regularization}
\label{Sec.advantage_forward_kl}
In fact, \citet{diffusionDPO} use the objective Eq.~\eqref{eq.general_obj_r_k_reg}, which employs reverse KL regularization to address OOD issues. They provide a different alignment loss as
\begin{equation}
\label{eq.loss_dpo_rk}
\begin{aligned}
    &\mathcal{L}_{\rm{DPORK}} \approx -\mathbb{E}_{(\boldsymbol{\sigma}^+,\boldsymbol{\sigma}^-)\sim\mathcal{D}_{\rm{pref}}}\mathbb{E}_{t\sim\mathcal{U}(1,T),\boldsymbol{\epsilon}_t^+,\boldsymbol{\epsilon}_t^-\sim\mathcal{N}(0,I)}\\&\rm{Sigmoid}\left(-\rho\left(\underbrace{\left\|\boldsymbol{\epsilon}_t^+ - \boldsymbol{\epsilon}_{\theta}(\boldsymbol{\sigma}_t^+,t)\right\|^2 - \left\|\boldsymbol{\epsilon}_t^- - \boldsymbol{\epsilon}_{\theta}(\boldsymbol{\sigma}_t^-,t)\right\|^2}_{\text{preference}} \right.\right.\\&\qquad\left.\left.- \underbrace{\left\|\boldsymbol{\epsilon}_t^+ - \boldsymbol{\epsilon}_{\rm{ref}}(\boldsymbol{\sigma}_t^+,t)\right\|^2 - \left\|\boldsymbol{\epsilon}_t^- - \boldsymbol{\epsilon}_{\rm{ref}}(\boldsymbol{\sigma}_t^-,t)\right\|^2}_{\text{regularization}} \right)\right).
\end{aligned}
\end{equation}
We can see that Eq.~\eqref{eq.loss_dpo_rk} contains the same preference term but a different regularization term. This regularization term does not involve trainable policy parameters $\pi_{\theta}$. It acts more like introducing a bias within the nonlinear Sigmoid function, rather than directly controlling the D-MSE of $\pi_{\theta}$ across the entire dataset as Eq.~\eqref{eq.loss_alignment}. Therefore, Eq.~\eqref{eq.loss_dpo_rk} might guide $\pi_{\theta}$ in an incorrect direction, where the D-MSE of winning segments is significantly lower than that of losing segments, while the overall D-MSE of both winning and losing segments increases. In other words, the likelihood of all segments may decrease, even though the winning segments have a higher likelihood than the losing segments. This leads to the resulting $\pi_{\theta}$ suffering from OOD issues. 

\begin{figure}
    \centering
    \includegraphics[width=1.\linewidth]{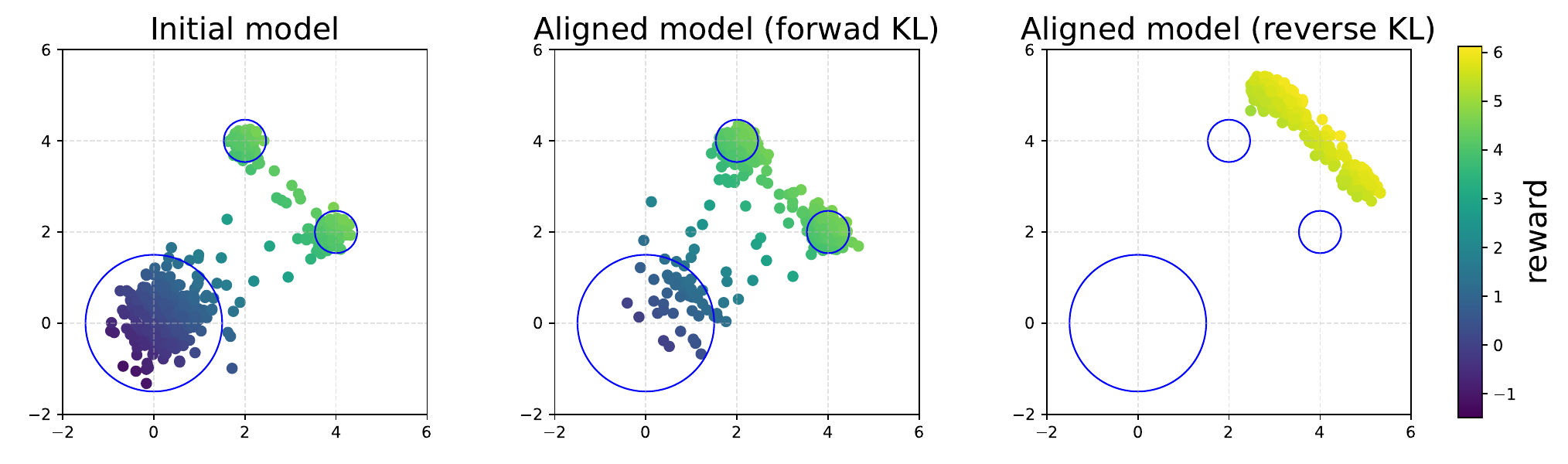}
    \caption{Samples generated by a toy diffusion model. The left sub-figure displays samples generated by the initial model. The middle and right sub-figures show samples generated by the aligned models with forward and reverse KL regularization, respectively. }
    \label{fig.toy_model}
\end{figure}

We also provide a toy model to intuitively demonstrate the distinction between reverse KL and forward KL regularization during the alignment process of diffusion policy. As shown in Fig.~\ref{fig.toy_model}, the initial diffusion policy (left) is trained on a dataset following a two-dimensional mixture Gaussian distribution. The centers and radii of blue circles represent the mean and standard deviation of Gaussian components.
The reward is defined as the dot product between a sample and the vector $(1,1)/\sqrt{2}$. We then build preference dataset using the script teacher~\cite{PT}, and align the initial policy with the preference dataset using forward KL and reverse KL regularization, respectively. The samples generated by the two aligned models are shown in the middle and right sub-figures of Fig.~\ref{fig.toy_model}. It is evident that the diffusion model aligned with reverse KL regularization generates a significant number of OOD samples, despite these samples having fairly high reward values. In contrast, the model aligned with forward KL regularization tends to produce samples with relatively high reward values, with most samples not exhibiting OOD issues. In conclusion, forward KL regularization is more effective in avoiding OOD issues during the alignment process of diffusion models.

\section{Related Work}
\paragraph{Diffusion Model for Decision Making.} Diffusion model is a kind of generative model that exhibits strong capabilities in generating high-dimensional samples \cite{song2020denoising,lu2022dpm}, e.g., images and text \cite{imagic,textdiffuser}. Diffusion models have been recently adopted in RL to solve various decision-making problems \cite{diffusion-survey}. Diffuser \cite{diffuser} first proposes to generate state-action trajectories via classifier-guided sampling. Decision-Diffusion \cite{decisiondiffuser}, MTDiff \cite{MTDiff}, MetaDiffuser \cite{metadiffuser} and SkillDiffuser \cite{skilldiffuser} extend such a framework by using various properties of trajectories for classifier-free guidance, such as task prompts, demonstrations, constraints, skills, representations, and attributes. Other works like Diffusion-QL \cite{DQL}, EDP \cite{kang2024efficient}, and IDQL \cite{IDQL} treat the conditional diffusion model as a policy and update the policy to maximize the value function. However, the above methods require manually designed reward functions for conditional generation, and learned policy cannot be aligned to preferences. For imitation learning, Diffusion Policy \cite{diffusionpolicy}, VLP \cite{VLP}, and UniPi \cite{UniPi} predict future states or actions based on expert demonstrations.
AlignDiff \cite{aligndiff} tries to align the diffusion model with preferences by first training an attribute strength model and then performing conditional generation for diffusion models. In contrast, we take an alternative way for preference alignment via policy optimization without requiring an attribute model for trajectory labeling. Other works also adopt the diffusion model as a data synthesizer \cite{SER}, environment model \cite{unisim}, and reward generator \cite{diffusion-reward} for decision-making, while their contributions are orthogonal to our work. 

\paragraph{RLHF and Preference-based RL.} RLHF has gained much attention in aligning LLMs with human preference \cite{RLHF-1,RLHF-2}. RLHF usually contains two stages that learn a reward function first \cite{BTmodel,rewardbench} and optimize the LLMs via RL algorithms \cite{li2023remax,ahmadian2024back}. Following this paradigm, Preference-based RL tries to align the policy output with preference data in decision-making problems by first learning the reward function through a BT model \cite{PT,bpref} and performing policy optimization through RL \cite{MRN}. Recently, direct preference learning (DPO) \cite{DPO} and the follow-up works \cite{rafailov2024r,wang2024beyond} proposed a novel perspective that implicitly optimizes the same objective as RLHF via a KL-constrained reward maximization framework, which forms an implicit reward function. Following this, DPPO \cite{DPPO} and Contrastive Preference Learning (CPL) \cite{CPL} extend direct preference learning to solve preference-based RL problems via contrastive learning of policy likelihoods or distances. Compared to these methods, we use an advanced diffusion model to enhance the expression ability of policies, while it causes further challenges in a DPO optimization, and we solve it via an approximate objective and additional regularization. The idea of preference alignment has recently been used for diffusion models to improve visual appeal and text alignment in image generation problems through two-stage RLHF \cite{diffusion-rl, DPOK} and direct preference optimization \cite{clark2023directly,diffusionDPO,yang2023using}. In contrast, our method focuses on learning diffusion policy through RLHF and proposes the first DPO-based learning framework that learns policies directly from preferences without RL. Regularization methods, particularly reverse KL and forward KL, have been widely discussed in the alignment of LLM~\cite{sun2024inverse}. Reverse KL and forward KL are known for their mode-seeking and mass-covering characteristics, respectively~\cite{wang2024beyond}. Moreover, forward KL is better suited for balancing alignment performance and generation diversity. In this paper, we highlight another significant advantage of forward KL in the alignment of diffusion policies, namely its enhanced effectiveness in avoiding OOD issues.

\section{Experiments}
In this section, we carry out extensive experiments to answer the following three questions. \textbf{Q1}:~Is FKPD an effective solver for the PBRL problem? \textbf{Q2}:~Is forward KL regularization crucial to FKPD, and is it more effective than reverse KL regularization? \textbf{Q3}:~How does FKPD perform under different hyperparameter settings? (Please refer to Appendix 2.1)

We evaluate the performance of FKPD on two well-known benchmarks: MetaWorld robotics tasks~\cite{metaworld} and D4RL locomotion~\cite{d4rl} tasks. The first benchmark uses the same offline dataset as Contrastive Preference Learning (CPL)~\cite{CPL}. This dataset is collected on six tasks from the simulated MetaWorld robotics environment~\cite{metaworld}. For each task, the preference-free dataset is collected by rolling out $2500$ episodes of length $250$ with a sub-optimal stochastic policy. The preference dataset is collected by uniformly sampling segments of length $64$ from the preference-free dataset and adding preference labels using script teacher~\cite{PT}. In terms of the D4RL benchmark, we generate the preference dataset from the original dataset using the same procedure as~\cite{CPL}. For further details about the dataset, please refer to Appendix 3.1.

For baseline methods, we choose three strong baselines. The first baseline is supervised fine-tuning (\textbf{SFT})~\cite{lee2018training}, where a policy is first trained with behavior clone on all segments in $\mathcal{D}_{\rm{pref}}$, then further fine-tuned on only the winning segments in $\mathcal{D}_{\rm {pref}}$. The second baseline is Preference-IQL (\textbf{P-IQL})~\cite{PT}, which learns a preference model from $\mathcal{D}_{\rm{pref}}$, then learns a policy to maximize it with Implicit $Q$-learning (IQL)~\cite{IQL}, an efficient offline RL algorithm. The third one is \textbf{CPL}, a state-of-the-art baseline that also uses DPO for the alignment phase while employs only a basic Gaussian policy. 

\begin{table}[h!]
  \begin{center}
    \setlength{\tabcolsep}{2.5pt}
    \begin{tabular}{lccccc}
    \toprule
    & & SFT & P-IQL & CPL & FKPD\\
    \hline
    \multirow{6}{*}{\rotatebox{90}{\makecell[c]{2.5k Dense}}} 
    &BinPicking     &66.9$\pm{2.1}$   &70.6$\pm{4.1}$   &\textbf{80.0}$\pm{2.5}$   &77.0$\pm{3.6}$\\
    &ButtonPress    &21.6$\pm{1.6}$   &16.2$\pm{5.4}$   & 24.5$\pm{2.1}$   & \textbf{34.7}$\pm{2.1}$\\
    &DoorOpen       &63.3$\pm{1.9}$   &69.0$\pm{6.2}$   & 80.0$\pm{6.8}$   & \textbf{87.7}$\pm{1.2}$\\
    &DrawerOpen     &62.6$\pm{2.4}$   &71.1$\pm{2.3}$   & 83.6$\pm{1.6}$   & \textbf{90.0}$\pm{2.7}$\\
    &PlateSlide     &41.6$\pm{3.5}$   &49.6$\pm{3.4}$   &\textbf{61.1}$\pm{3.0}$   &58.3$\pm{1.2}$\\
    &SweepInto      &51.9$\pm{2.1}$   &60.6$\pm{3.6}$   &70.4$\pm{3.0}$    &\textbf{76.5}$\pm{3.5}$\\
    \hline
    \multirow{6}{*}{\rotatebox{90}{\makecell[c]{20k Sparse}}}
    &BinPicking     &67.0$\pm{4.9}$   &75.0$\pm{3.3}$   &\textbf{83.2}$\pm{3.5}$   &76.3$\pm{1.5}$\\
    &ButtonPress    &21.4$\pm{2.7}$   &19.5$\pm{1.8}$   &29.8$\pm{1.8}$   &\textbf{33.3}$\pm{4.5}$\\
    &DoorOpen       &63.6$\pm{2.4}$   &79.0$\pm{6.6}$   &77.9$\pm{5.0}$   &\textbf{84.3}$\pm{2.1}$\\
    &DrawerOpen     &63.5$\pm{0.9}$   &76.2$\pm{2.8}$   &79.1$\pm{5.0}$   &\textbf{91.3}$\pm{2.3}$\\
    &PlateSlide     &41.9$\pm{3.1}$   &55.5$\pm{4.2}$   &56.4$\pm{3.9}$   &\textbf{66.3}$\pm{7.4}$\\
    &SweepInto      &50.9$\pm{3.2}$   &73.4$\pm{4.2}$   &\textbf{81.2}$\pm{1.6}$   &76.3$\pm{2.1}$\\
    \bottomrule
    \end{tabular}
    \caption{Success rates (in percent) of all methods for six MetaWorld tasks on different datasets. Here $2.5$k Dense and $20$k sparse represent two different preference datasets, with detailed explanations provided in Appendix 3.1.} 
    \label{tab.table11}
  \end{center}
\end{table}
\vspace{-2pt}

\begin{table}
    \begin{center}
    \setlength{\tabcolsep}{2.5pt}
    \setlength{\tabcolsep}{3pt}
        \begin{tabular}{llcccc}
            \toprule
            \small
                           & & SFT  & P-IQL             &CPL             &FKPD          \\
            \hline
            \multirow{3}{*}{\rotatebox{90}{\makecell[c]{expert}}} 
             &walk2d&  82.9$\pm{8.5}$&\textbf{109.8}$\pm{0.4}$ & 107.9$\pm{0.2}$ & 107.8$\pm{0.2}$\\
             &hopper&  48.3$\pm{5.0}$&\textbf{84.5}$\pm{4.1}$ & 64.5$\pm{6.9}$ & 80.5$\pm{3.4}$\\
             
             &halfchee&  68.5$\pm{4.7}$&83.6$\pm{3.8}$ & 90.9$\pm{2.3}$ & \textbf{94.7}$\pm{0.3}$\\

            \hline
            \multirow{3}{*}{\rotatebox{90}{\makecell[c]{replay}}} 
             &walk2d&  33.9$\pm{7.1}$&\textbf{71.2}$\pm{10.3}$ & 48.3$\pm{3.7}$ & 69.3$\pm{3.8}$\\
             
             &hopper&  57.6$\pm{5.6}$&68.9$\pm{33.8}$ & \textbf{111.2}$\pm{0.2}$ & 100.7$\pm{1.6}$\\
             
             &halfchee&  34.8$\pm{1.8}$&42.3$\pm{0.5}$ & 45.3$\pm{0.1}$ & \textbf{45.5}$\pm{0.2}$\\ 
            \bottomrule
        \end{tabular}
        \caption{Average norm score of all methods for D4RL locomotion tasks on different datasets.}
        \label{tab.table2}
    \end{center}
\end{table}

\subsection{Performance of FKPD}
The results of FKPD and competitors for MetaWorld are presented in Table \ref{tab.table11}. The architecture of the policy network, along with evaluation details and hyperparameter settings, are provided in Appendix 3-4. When we use the dense comparison data, FKPD achieves the best performance in $4$ of $6$ tasks. Especially in \textit{Button Press}, \textit{Door Open}, and \textit{Drawer Open}, FKPD has significant advantages over its competitors. When using sparser comparison data, FKPD also outperforms its competitors in $4$ of $6$ tasks, with a substantial margin in \textit{Button Press}, \textit{Door open} and \textit{Drawer Open}. In conclusion, FKPD is the method with the best overall performance in this dataset. Compared to SFT, FKPD makes more effective use of the information in $\mathcal{D}_{\rm pref}$, leading to better performance.
Compared with P-IQL, FKPD directly optimizes the policy using preference data, without learning an explicit preference score function. This eliminates the risk associated with uncertain score functions. Compared with CPL, FKPD adopts a more expressive diffusion policy, making it more suitable for complex manipulation tasks.

The results of FKPD and competitors for D4RL are demonstrated in Table \ref{tab.table2}. For this benchmark, we also use a critic as~\cite{DQL}, which is directly learned from the preference dataset~\cite{hejna2024inverse} without the need of a reward model, to filter out sub-optimal candidate actions. We observe that the proposed FKPD outperforms CPL in $5$ out of $6$ D4RL locomotion tasks, further demonstrating the advantages of employing a diffusion strategy. Additionally, it achieves competitive performance with the two-stage P-IQL without the need of training an explicit reward model. In a word, FKPD is an effective solver for D4RL locomotion tasks. 

\subsection{Ablation of Regularization}
We next provide a detailed comparison of two aforementioned regularization methods: forward KL and reverse KL regularization. To this end, we compare FKPD and Reverse KL regularized for aligning Diffusion Policies (RKPD)~\cite{diffusionDPO} on three representative MetaWorld ($20$k sparse) tasks, namely, \textit{Door Open}, \textit{Drawer Open} and \textit{Button Press}; as well as three representative D4RL locomotion tasks, namely, \textit{Walker2d-medium-replay}, \textit{hopper-medium-replay} and \textit{halfcheetah-medium-replay}. Moreover, we also provide the results of DPO without regularization (referred to as NRPD) for reference. Here we use a novel metric to evaluate the performance of the alignment phase for these three methods. For each method, we record the policy's average success rate (or average return for D4RL tasks) on a task as $U_0$ at the beginning of the alignment phase. After the alignment phase, we evaluate its average success rate (or average return) again, recorded as $U_1$. Then, we define the improvement factor as $F_{\rm im}=(U_1-U_0)/U_0$. All results are present in Table~\ref{tab.table3}.

For three MetaWorld tasks, we observe that NRPD exhibits the worst performance among the three tasks, especially in \textit{Button Press}, where it completely fails. This demonstrates the importance of regularization terms in diffusion policy learning. In addition, FKPD outperforms RKPD in the \textit{Door Open} and \textit{Drawer Open} tasks, performs slightly worse in the Button Press task, and overall demonstrates better performance. For three D4RL tasks, We observe that FKPD demonstrates a more pronounced advantage. At this point, RKPD and NRPD can not even enhance policy performance during the alignment phase. This implies that regularization is is of greater importance in D4RL tasks.
\begin{table}
    \begin{center}
    \setlength{\tabcolsep}{3pt}
        \begin{tabular}{cccc}
            \toprule
            \small
                           &FKPD             &RKPD             &NRPD          \\
            \hline
             Door Open&  \textbf{40.0}$\pm{3.5}$ & 21.7$\pm{8.8}$ & -42.1$\pm{25.6}$\\
             Drawer Open&  \textbf{47.2}$\pm{3.7}$ & 6.9$\pm{3.4}$ & -73.7$\pm{24.5}$\\
             Button Press&  58.6$\pm{21.6}$ & \textbf{60.9}$\pm{21.0}$ & -100.0$\pm{0.0}$\\
            \hline
             Walk2d-med-rep&  \textbf{75.4}$\pm{5.0}$ & -88.7$\pm{11.9}$ & -95.1$\pm{6.8}$\\
             hopper-med-rep&  \textbf{64.8}$\pm{12.5}$ & -33.9$\pm{17.9}$ & -98.4$\pm{1.0}$\\
             halfchee-med-rep&  \textbf{14.9}$\pm{1.2}$ & -45.3$\pm{54.3}$ & -100.6$\pm{9.4}$\\
            \bottomrule
        \end{tabular}
        \caption{Average improvement factor (in percentage) of three DPO methods on six MetaWorld and D4RL tasks.}
        \label{tab.table3}
    \end{center}
\end{table}

Next we conduct an in-depth analysis of why forward KL regularization outperforms reverse KL regularization for the aligning of diffusion policy. To this end, we track three variables in the process of alignment. The first two variables are average D-MSE of winning segments and losing segments, abbreviated as $E_{\rm winning}$ and $E_{\rm losing}$, respectively.

\begin{figure}[htp]
    \centering
    \includegraphics[width=1.05 \linewidth]{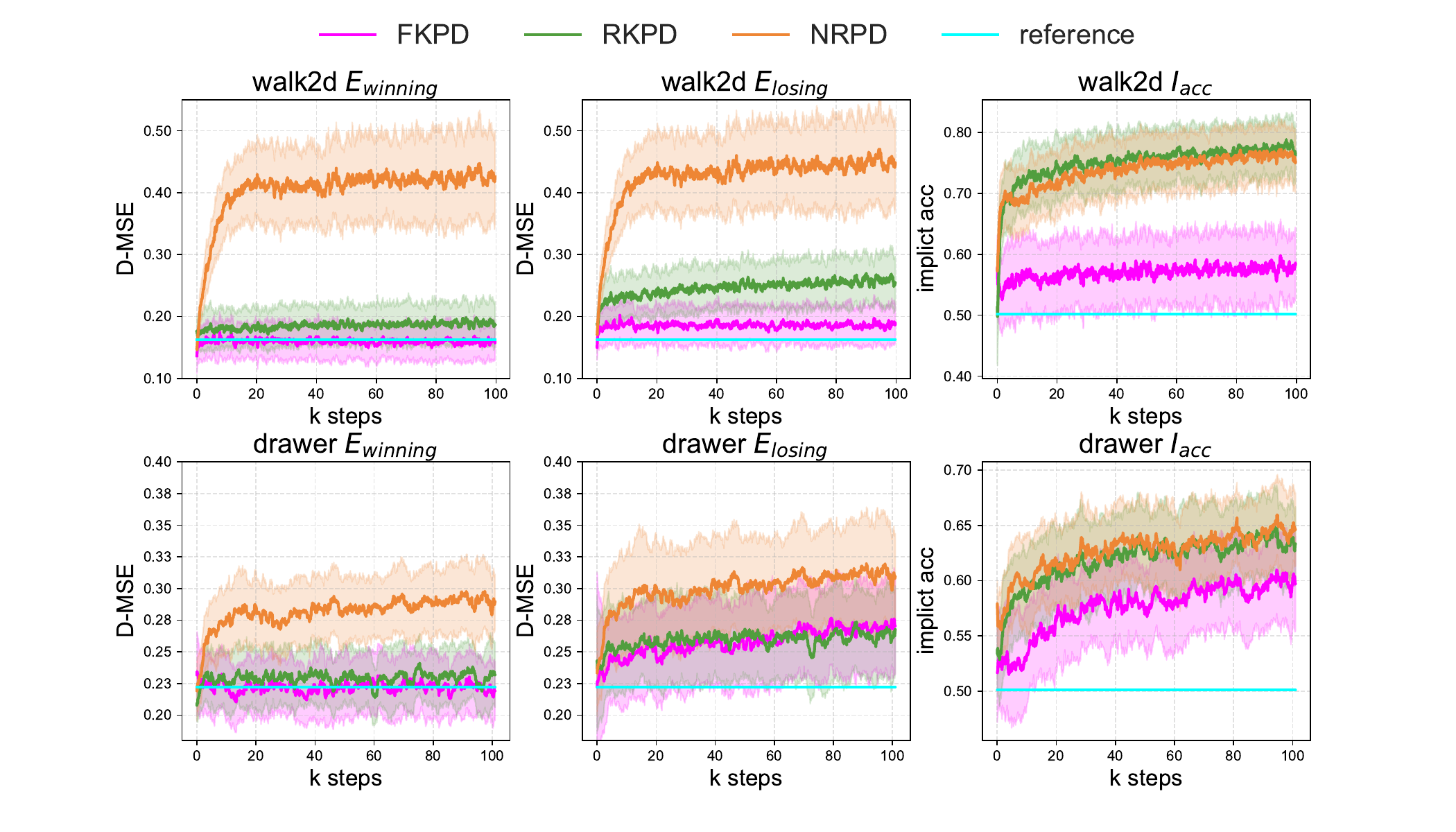}
    \caption{Average D-MSE and implicit accuracy for FKPD, NKPD, and NRPD during the alignment phase of \textit{Drawer Open} (bottom row) and \textit{Walker2d-medium-replay} (top row). We also provide the same variables of the pretrained policy on the preference-free dataset $\mathcal{D}$ for reference.}
    \label{fig.ablation_regularization}
\end{figure}

They are capable of measuring the variation in the likelihood of winning segments and losing segments during the alignment process of diffusion policy. The third variable is implicit accuracy, which is written as
\begin{equation}
    I_{\rm acc}\!=\!\frac{1}{B}\!\!\!\!\!\!\!\sum_{\boldsymbol{\sigma^+}\boldsymbol{\sigma^-}\in\mathcal{B}_{\rm pref}}\!\!\!\!\!\!\!\!\!\!\mathbb{I}\left(\left\|\boldsymbol{\epsilon}_t^+ \!\!\!-\!\! \boldsymbol{\epsilon}_{\theta}(\boldsymbol{\sigma}_t^+,t)\right\|^2 \!\!\!- \left\|\boldsymbol{\epsilon}_t^- - \boldsymbol{\epsilon}_{\theta}(\boldsymbol{\sigma}_t^-,t)\right\|^2\!\!<0\right),
\end{equation}
where $\mathbb{I}$ is the indicator function, $\mathcal{B}_{\rm pref}$ is a batch of segments with preference labels, $B$ is the size of $\mathcal{B}_{\rm pref}$. $I_{\rm acc}$ represents the estimated probability that the likelihood of $\pi_{\theta}$ generating wining segments is greater than that of generating losing segments.
Fig.~\ref{fig.ablation_regularization} illustrates these three variables for FKPD, RKPD, and NRPD in \textit{Drawer Open} and \textit{Walker2d-medium-replay}. We also provide these variables of the pre-trained policy from the behavior clone phase for reference. We start with NRPD. Although it ensures that the likelihood of generating winning segments increases relative to the likelihood of generating losing segments (as indicated by $I_{\rm acc}$ values) during training, the  $E_{\rm winning}$ and $E_{\rm losing}$ values reveal that the likelihood of both positive and negative segments decreases significantly. This suggests that the NRPD method suffers from a severe OOD issue. FKPD, while ensuring an increased likelihood of generating positive segments relative to negative segments, also effectively maintains the values of $E_{\rm winning}$ and $E_{\rm losing}$ without significant increases. This allows it to preserve the overall likelihood of both positive and negative segments, thereby addressing the OOD issue present in NRPD. RKPD, similar to NRPD, significantly increases the likelihood of positive segments relative to negative segments. However, the likelihood for both the positive and negative segments decreases noticeably compared to the reference values. Thus, it fails to address the OOD challenge as effectively as the FKPD.

\section{Conclusions}
In this paper, we propose FKPD, a method to align a diffusion policy with preference data. To tackle the OOD issue during the alignment phase, we employ the forward KL regularization, which demonstrates superior performance compared to the reverse KL regularization in the experiments. FKPD achieves the sate-of-the-art performance on open MetaWorld and D4RL locomotion tasks, indicating that it is an efficient preference aligning method. 
Currently, we have only validated the effectiveness of FKPD for single-modal observations, that is, the joint states of embodiment, and single-task scenarios. In future work, we aim to extend FKPD to encompass multi-task and multi-modal data. Moreover, other efficient regularization methods, such as those proposed in \cite{huang2024correcting, cen2024value}, have been introduced for aligning LLMs. In our future work, we plan to analyze the performance of these new regularization methods for aligning diffusion policies. 

\section{Acknowledgments}
This study is supported by National Natural Science Foundation of China (Grant No.62306242). Zhao extends gratitude to professor Chenjia for providing in-depth guidance on this work. Thanks are also extended to Chenyou and the other authors for their outstanding contributions to the research. 

\bibliography{main}
\clearpage


\section*{Supplementary material}

# Appendix of FKPD

## 1 Theoretical Proof

### 1.1 Optimal generative policy with respect to preference score function

The optimization problem for the optimal generative policy of the maximum entropy reinforcement learning is as follows:

$$\max_{\pi_\theta} \int \left( r(\boldsymbol{\sigma}) - \rho \log \pi_\theta(\boldsymbol{\sigma}) \right) d\boldsymbol{\sigma}$$
$$\text{s.t.} \int \pi_\theta(\boldsymbol{\sigma}) d\boldsymbol{\sigma} = 1 \tag{1}$$

Next, we form the Lagrangian:

$$\mathcal{L}(\pi_\theta, \rho, \eta) = \int \left( r(\boldsymbol{\sigma}) - \rho \log \pi_\theta(\boldsymbol{\sigma}) \right) d\boldsymbol{\sigma} - \eta \left( \int \pi_\theta(\boldsymbol{\sigma}) d\boldsymbol{\sigma} - 1 \right), \tag{2}$$

where $\eta$ is a Lagrange multiplier. Differentiating $\mathcal{L}(\pi_\theta, \rho, \eta)$ with respect to $\pi_\theta(\boldsymbol{\sigma})$ results in:

$$\frac{\partial \mathcal{L}}{\pi_\theta(\boldsymbol{\sigma})} = r(\boldsymbol{\sigma}) - \rho \log \pi_\theta(\boldsymbol{\sigma}) - \eta. \tag{3}$$

Setting to zero and solving for $\pi_\theta(\boldsymbol{\sigma})$ gives

$$\pi_\theta^*(\boldsymbol{\sigma}) = \frac{1}{Z} \exp\left( \frac{r(\boldsymbol{\sigma})}{\rho} \right), \tag{4}$$

where $Z = \exp(\frac{\eta}{\rho})$ is the partition function.

### 1.2 Preference model for diffusion policy

Define the score function on the whole reveres chain as $R(\boldsymbol{\sigma}_{0:T})$. Then the original score function $r$ can be calculated by marginalizing out all mediating variables as :

$$r(\boldsymbol{\sigma}_0) = \mathbb{E}_{\boldsymbol{\sigma}_{1:T} \sim \pi_\theta(\boldsymbol{\sigma}_{1:T}|\boldsymbol{\sigma}_0)} R(\boldsymbol{\sigma}_{0:T}). \tag{5}$$

In addition, we can also write the objective for the whole reverse process as:

$$\max_{\pi_\theta} \mathbb{E}_{\boldsymbol{\sigma}_{0:T} \sim \pi_\theta} \left[ R(\boldsymbol{\sigma}_{0:T}) - \rho \log \pi_\theta(\boldsymbol{\sigma}_{0:T}) \right]. \tag{6}$$

After solving for Eq. (6), we have the relationship between the optimal diffusion policy and the preference score function on the whole reverse chain as:

$$\pi_\theta^*(\boldsymbol{\sigma}_{0:T}) = \frac{1}{Z} \exp\left( \frac{R(\boldsymbol{\sigma}_{0:T})}{\rho} \right). \tag{7}$$

Then we substitute Eq. (5) (7) to the Bradley-Terry model, obtaining the preference model for a diffusion policy as:

$$P(\boldsymbol{\sigma}^+ \succ \boldsymbol{\sigma}^-) = \text{Sigmoid}(r(\boldsymbol{\sigma}_0^+) - r(\boldsymbol{\sigma}_0^-))$$
$$= \text{Sigmoid}\left( \rho \left( \mathbb{E}_{\boldsymbol{\sigma}_{1:T}^+ \sim \pi_\theta^*(\boldsymbol{\sigma}_{1:T}^+|\boldsymbol{\sigma}_0^+)} \log \pi_\theta^*(\boldsymbol{\sigma}_{0:T}^+) - \mathbb{E}_{\boldsymbol{\sigma}_{1:T}^- \sim \pi_\theta^*(\boldsymbol{\sigma}_{1:T}^-|\boldsymbol{\sigma}_0^-)} \log \pi_\theta^*(\boldsymbol{\sigma}_{0:T}^-) \right) \right). \tag{8}$$

Subsequently, we utilize contrastive learning, obtaining the DPO objective for a diffusion policy as:

$$\mathcal{L}_{\text{DPODiff}}(\pi_\theta, \mathcal{D}_{\text{pref}}) = \mathbb{E}_{(\boldsymbol{\sigma}^+, \boldsymbol{\sigma}^-) \sim \mathcal{D}_{\text{pref}}} \left[ -\text{F}\left( \mathbb{E}_{\boldsymbol{\sigma}_{1:T}^+ \sim \pi_\theta(\boldsymbol{\sigma}_{1:T}^+|\boldsymbol{\sigma}_0^+)} \log \pi_\theta(\boldsymbol{\sigma}_{0:T}^+) - \mathbb{E}_{\boldsymbol{\sigma}_{1:T}^- \sim \pi_\theta(\boldsymbol{\sigma}_{1:T}^-|\boldsymbol{\sigma}_0^-)} \log \pi_\theta(\boldsymbol{\sigma}_{0:T}^-) \right) \right]. \tag{9}$$

### 1.3 Simplification of DPO objective for Diffusion policy

The original DPO objective is written as:

$$\mathcal{L}_{\text{DPODiff}}(\pi_\theta, \mathcal{D}_{\text{pref}}) = \mathbb{E}_{(\boldsymbol{\sigma}^+, \boldsymbol{\sigma}^-) \sim \mathcal{D}_{\text{pref}}} \left[ -F \left( \mathbb{E}_{\substack{\boldsymbol{\sigma}^+_{1:T} \sim \pi_\theta(\boldsymbol{\sigma}^+_{1:T}|\boldsymbol{\sigma}^+_0) \\ \boldsymbol{\sigma}^-_{1:T} \sim \pi_\theta(\boldsymbol{\sigma}^-_{1:T}|\boldsymbol{\sigma}^-_0)}} \left[ \log \pi_\theta(\boldsymbol{\sigma}^+_{0:T}) - \log \pi_\theta(\boldsymbol{\sigma}^-_{0:T}) \right] \right) \right]. \tag{10}$$

Since sampling from $\pi_\theta(\boldsymbol{\sigma}_{0:T})$ is intractable, we utilize the forward process $q(\boldsymbol{\sigma}_{1:T}|\boldsymbol{\sigma}_0)$ to approximate the reverse process, yielding the following objective:

$$\mathcal{L}_{\text{DPODiff}}(\pi_\theta, \mathcal{D}_{\text{pref}}) \approx \mathcal{L}_1(\pi_\theta, \mathcal{D}_{\text{pref}})$$
$$= \mathbb{E}_{(\boldsymbol{\sigma}^+, \boldsymbol{\sigma}^-) \sim \mathcal{D}_{\text{pref}}} \left[ -F \left( \mathbb{E}_{\substack{\boldsymbol{\sigma}^+_{1:T} \sim q(\boldsymbol{\sigma}^+_{1:T}|\boldsymbol{\sigma}^+_0) \\ \boldsymbol{\sigma}^-_{1:T} \sim q(\boldsymbol{\sigma}^-_{1:T}|\boldsymbol{\sigma}^-_0)}} \left[ \log \pi_\theta(\boldsymbol{\sigma}^+_{0:T}) - \log \pi_\theta(\boldsymbol{\sigma}^-_{0:T}) \right] \right) \right]. \tag{11}$$

We next focus on the contents inside function $F(\cdot)$, denote

$$C = \mathbb{E}_{\boldsymbol{\sigma}^+_{1:T} \sim q(\boldsymbol{\sigma}^+_{1:T}|\boldsymbol{\sigma}^+_0), \boldsymbol{\sigma}^-_{1:T} \sim q(\boldsymbol{\sigma}^-_{1:T}|\boldsymbol{\sigma}^-_0)} \left[ \log \pi_\theta(\boldsymbol{\sigma}^+_{0:T}) - \log \pi_\theta(\boldsymbol{\sigma}^-_{0:T}) \right]. \tag{12}$$

Subsequently, we decompose the reverse process in Eq. (12) as $\pi_\theta(\boldsymbol{\sigma}) = \pi_\theta(\boldsymbol{\sigma}_T) \prod_{t=1}^T \pi_\theta(\boldsymbol{\sigma}_{t-1}|\boldsymbol{\sigma}_t)$, and perform the following algebraic operations:

$$\begin{aligned}
C &= \mathbb{E}_{\boldsymbol{\sigma}^+_{1:T}, \boldsymbol{\sigma}^-_{1:T}} \left[ \log \pi_\theta(\boldsymbol{\sigma}^+_{0:T}) - \log \pi_\theta(\boldsymbol{\sigma}^-_{0:T}) \right] \\
&= \mathbb{E}_{\boldsymbol{\sigma}^+_{1:T}, \boldsymbol{\sigma}^-_{1:T}} \left[ \sum_{t=1}^T \log \pi_\theta(\boldsymbol{\sigma}^+_{t-1}|\boldsymbol{\sigma}^+_t) - \sum_{t=1}^T \log \pi_\theta(\boldsymbol{\sigma}^-_{t-1}|\boldsymbol{\sigma}^-_t) \right] \\
&= \mathbb{E}_{\boldsymbol{\sigma}^+_{1:T}, \boldsymbol{\sigma}^-_{1:T}} T \mathbb{E}_t \left[ \log \pi_\theta(\boldsymbol{\sigma}^+_{t-1}|\boldsymbol{\sigma}^+_t) - \log \pi_\theta(\boldsymbol{\sigma}^-_{t-1}|\boldsymbol{\sigma}^-_t) \right] \\
&= T \mathbb{E}_t \mathbb{E}_{\boldsymbol{\sigma}^+_{t-1,t} \sim q(\boldsymbol{\sigma}^+_{t-1,t}|\boldsymbol{\sigma}^+_0), \boldsymbol{\sigma}^-_{t-1,t} \sim q(\boldsymbol{\sigma}^-_{t-1,t}|\boldsymbol{\sigma}^-_0)} \left[ \log \pi_\theta(\boldsymbol{\sigma}^+_{t-1}|\boldsymbol{\sigma}^+_t) - \log \pi_\theta(\boldsymbol{\sigma}^-_{t-1}|\boldsymbol{\sigma}^-_t) \right] \\
&= T \mathbb{E}_t \mathbb{E}_{\boldsymbol{\sigma}^+_t \sim q(\boldsymbol{\sigma}^+_t|\boldsymbol{\sigma}^+_0), \boldsymbol{\sigma}^-_t \sim q(\boldsymbol{\sigma}^-_t|\boldsymbol{\sigma}^-_0)} \\
&\quad \mathbb{E}_{\boldsymbol{\sigma}^+_{t-1} \sim q(\boldsymbol{\sigma}^+_{t-1}|\boldsymbol{\sigma}^+_0, \boldsymbol{\sigma}^+_t), \boldsymbol{\sigma}^-_{t-1} \sim q(\boldsymbol{\sigma}^-_{t-1}|\boldsymbol{\sigma}^-_0, \boldsymbol{\sigma}^-_t)} \left[ \log \pi_\theta(\boldsymbol{\sigma}^+_{t-1}|\boldsymbol{\sigma}^+_t) - \log \pi_\theta(\boldsymbol{\sigma}^-_{t-1}|\boldsymbol{\sigma}^-_t) \right]
\end{aligned} \tag{13}$$

Noting that $F(\cdot)$ is a convex function, we can take the expectations of $t$ and $\boldsymbol{\sigma}_t$ out of $F(\cdot)$, yielding:

$$\mathcal{L}_1(\pi_\theta, \mathcal{D}_{\text{pref}}) \leq \mathbb{E}_{(\boldsymbol{\sigma}^+, \boldsymbol{\sigma}^-) \sim \mathcal{D}_{\text{pref}}} \mathbb{E}_{t, \boldsymbol{\sigma}^+_t \sim q(\boldsymbol{\sigma}^+_t|\boldsymbol{\sigma}^+_0), \boldsymbol{\sigma}^-_t \sim q(\boldsymbol{\sigma}^-_t|\boldsymbol{\sigma}^-_0)}$$
$$\left[ -F \left( T \mathbb{E}_{\boldsymbol{\sigma}^+_{t-1} \sim q(\boldsymbol{\sigma}^+_{t-1}|\boldsymbol{\sigma}^+_0, \boldsymbol{\sigma}^+_t), \boldsymbol{\sigma}^-_{t-1} \sim q(\boldsymbol{\sigma}^-_{t-1}|\boldsymbol{\sigma}^-_0, \boldsymbol{\sigma}^-_t)} \left[ \log \pi_\theta(\boldsymbol{\sigma}^+_{t-1}|\boldsymbol{\sigma}^+_t) - \log \pi_\theta(\boldsymbol{\sigma}^-_{t-1}|\boldsymbol{\sigma}^-_t) \right] \right) \right]$$
$$= \mathbb{E}_{(\boldsymbol{\sigma}^+, \boldsymbol{\sigma}^-) \sim \mathcal{D}_{\text{pref}}} \mathbb{E}_{t, \boldsymbol{\sigma}^+_t \sim q(\boldsymbol{\sigma}^+_t|\boldsymbol{\sigma}^+_0), \boldsymbol{\sigma}^-_t \sim q(\boldsymbol{\sigma}^-_t|\boldsymbol{\sigma}^-_0)}$$
$$\left[ -F \left( -T \left( \mathbb{D}_{\text{kl}} \left( q(\boldsymbol{\sigma}^+_{t-1}|\boldsymbol{\sigma}^+_{0,t}) \| \pi_\theta(\boldsymbol{\sigma}^+_{t-1}|\boldsymbol{\sigma}^+_t) \right) - \mathbb{D}_{\text{kl}} \left( q(\boldsymbol{\sigma}^-_{t-1}|\boldsymbol{\sigma}^-_{0,t}) \| \pi_\theta(\boldsymbol{\sigma}^-_{t-1}|\boldsymbol{\sigma}^-_t) \right) \right) \right) \right]. \tag{14}$$

Finally, using the Gaussian parameterization of the reverse process, the above loss simplifies to:

$$\mathcal{L}_1(\pi_\theta, \mathcal{D}_{\text{pref}}) \leq \mathcal{L}_2(\pi_\theta, \mathcal{D}_{\text{pref}}) = -\mathbb{E}_{(\boldsymbol{\sigma}^+, \boldsymbol{\sigma}^-) \sim \mathcal{D}_{\text{pref}}} \mathbb{E}_{t \sim \mathcal{U}(1,T), \boldsymbol{\epsilon}^+_t, \boldsymbol{\epsilon}^-_t \sim \mathcal{N}(0,I)}$$
$$F \left( -T \left( \left\| \boldsymbol{\epsilon}^+_t - \boldsymbol{\epsilon}_\theta(\boldsymbol{\sigma}^+_t, t) \right\|^2 - \left\| \boldsymbol{\epsilon}^-_t - \boldsymbol{\epsilon}_\theta(\boldsymbol{\sigma}^-_t, t) \right\|^2 \right) \right), \tag{15}$$

where $\boldsymbol{\sigma}_t = \sqrt{\bar{\alpha}_t} \boldsymbol{\sigma}_0 + \sqrt{1 - \bar{\alpha}_t} \boldsymbol{\epsilon}$. It is obvious that the constant $T$ can be factored into $\rho$.

## 2 Additional Experimental Results

### 2.1 Ablation of hyperparameters

Finally, we ablate FKPD's hyperparameters on *Door Open* task with sparse 20k dataset. We consider three hyperparameters in our loss function: regularization weight $\mu$, bias $b$ and temperature $\rho$. Results are shown in Table 1. We can see that FKPD are not sensitive to temperature $\rho$, while it exhibits a slightly higher sensitivity to bias $b$ and the regularization weight $\mu$. However, in our experiments, we find that using a single set of hyperparameters can achieve good performance across most tasks. In conclusion, FKPD is a robust method that does not require extensive parameter tuning. This further demonstrates that FDPD is an effective method to align a diffusion policy with human preference.

| Hyperparameter | Value | Success rate |
|---|---|---|
| Regularization weight $\mu$ | 1.0 | **84.3**±2.1 |
| | 1.25 | 76.7 ± 3.8 |
| | 0.75 | 77.7 ± 13.6 |
| Bias $b$ | 0.0 | **84.3**±2.1 |
| | 0.25 | 78.3 ± 3.5 |
| | -0.25 | 80.0 ± 3.6 |
| Temperature $\rho$ | 500 | **84.3**±2.1 |
| | 5000 | 81.7 ± 1.5 |
| | 10 | 79.3 ± 1.5 |

Table 1: Ablation about key hyperparameters: regularization weight $\mu$, bias $b$ and temperature $\rho$.

## 2.2 Additional results on D4RL benchmark

We also carry out experiments on D4RL Adroit and Kitchen tasks. Due to space constraints in the main text, we present these experimental results in Tab. 2. The results of competitors are borrowed from [1]. We observe that FKPD outperforms CPL in Kitchen tasks, while it underperforms in Adroit tasks. Overall, FKPD and CPL exhibit comparable performance, with both methods significantly surpassing the PT-IQL approach.

| | | P-IQL | CPL | FKPD |
|---|---|---|---|---|
| Adroit | pen-human | 53.0±31.7 | **76.3**±14.4 | 53.1±4.5 |
| | pen-cloned | 42.9±24.4 | **75.1**±7.7 | 56.5±5.7 |
| kitchen | kitchen-mixed | 48.0±11.9 | 52.5±3.1 | **56.0**±3.5 |
| | kitchen-partial | 40.2±12.3 | 49.4±5.7 | **63.5**±3.3 |

Table 2: Average norm score of all methods for D4RL Adroit and kitchen tasks on different datasets.

## 2.3 Learning curves

In this section, we present the learning curves from our experiments:

- Fig. 1: Learning curves for MetaWorld tasks on dense datasets.
- Fig. 2: Learning curves for for MetaWorld tasks on sparse datasets.
- Fig. 3: Learning curves for regularization ablations in MetaWorld tasks.
- Fig. 4: Learning curves for regularization ablations in D4RL tasks.

## 3 Experimental Details

In this section, we provide experimental details of FKPD.

### 3.1 Dataset

We use the dataset from CPL [3], which is collected by a pre-trained suboptimal SAC agent. This pre-trained SAC agent can achieve approximately a 50% success rate and is used to gather episodes for all MetaWorld tasks. The collected episodes are then utilized to generate synthetic preference labels. Specifically, each episode is sampled into several segments of length 64. Another pre-trained

3

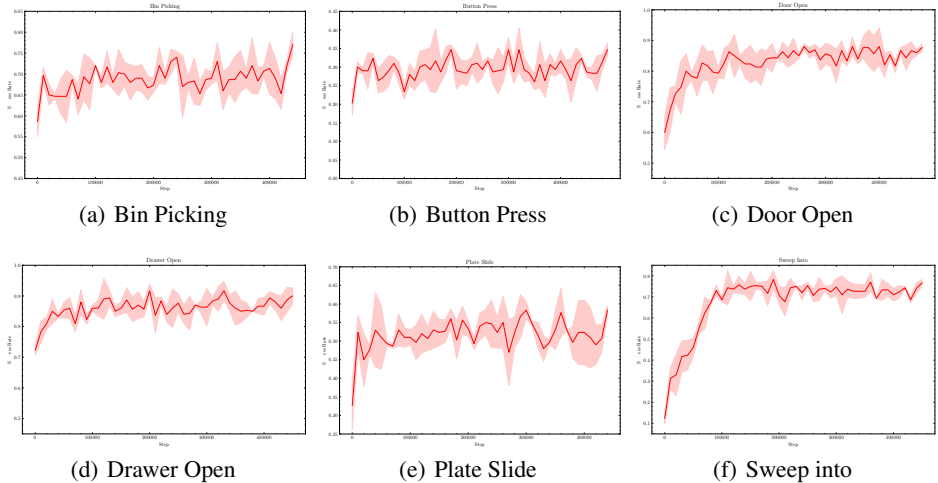

Figure 1: Success rate in MetaWorld with 2.5K segments and dense comparisons.

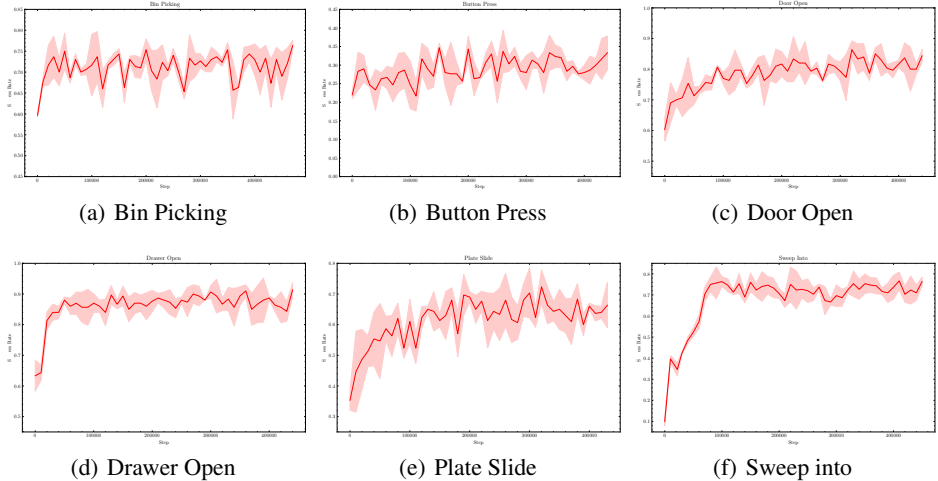

Figure 2: Success rate in MetaWorld with 20K segments and sparse comparisons.

SAC agent is then used to estimate the regret value of each segment. Based on the estimated regret value, segment pairs are compared to generate preference labels. There are two comparison methods corresponding to two dataset settings: *dense* and *sparse*. For the *dense* setting, comparisons occur between every sampled segment. For the *sparse* setting, each segment is compared only once. 2.5k dense means that the preference dataset contains 2.5k segments, and the segments are used in a dense comparison format. 20k sparse means that the preference dataset contains 20k segments, and the segments are used in a sparse comparison format. For D4RL benchmark, we collect similar dense comparison datasets for each task as CPL.

### 3.2 Evaluation

For MetaWorld benchmark, we report the final results after sufficient alignment steps. To draw a learning curve, we run 200 evaluation episodes and compute the average the success rate every 5000 steps during alignment, . In terms of the D4RL benchmark, we follow the same procedure as MetaWorld benchmark, except that we compute and report the average sum reward during the evaluation. For all results in our experiment, we report the mean and the standard error over 3 random seeds after 400k steps for all tasks.

4

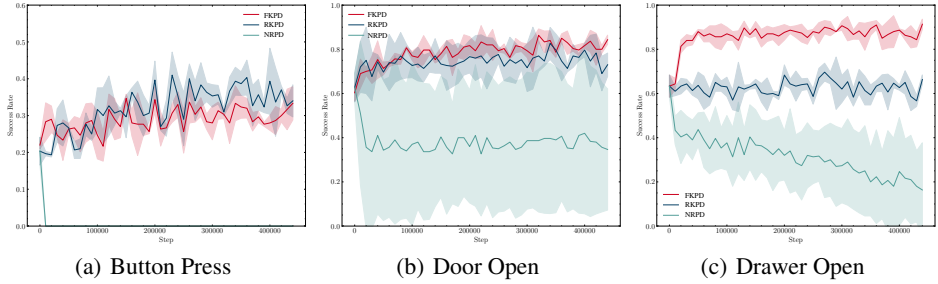

(a) Button Press    (b) Door Open    (c) Drawer Open

Figure 3: Success rate in MetaWorld for different regularization methods.

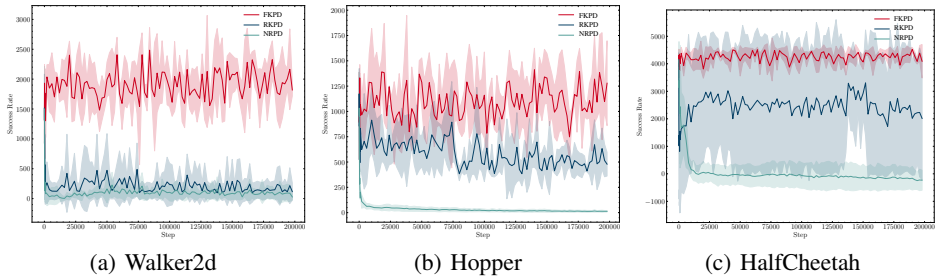

(a) Walker2d    (b) Hopper    (c) HalfCheetah

Figure 4: Average return in D4RL for different regularization methods.

## 4 Implementation Details

In this section, we provide implementation details of FKPD. For CPL method, we use the default hyperparameters recommended in [3].

### 4.1 Diffusion model

- We represent the noise model as a residual network with layer normalization [2]. The segment $\sigma$ is embedded into hidden features by a dense layer before being fed into $n$ MLPResNet Blocks. After the repeated blocks, a dense layer is used to reduce the output to the action dimension.
- Each MLPResNet Block consists of a dropout layer followed by layer normalization. This is succeeded by a dense layer with an output dimension of four times the hidden dimension, followed by an activation function, and a second dense layer that reduces the output dimension back to the hidden dimension. The input is directly added to the output of the second dense layer, forming a residual connection.
- We use $n = 2$ MLPResNet Blocks with hidden dim $256$.
- We use $T = 50$ for diffusion steps.

### 4.2 Hyperparameters

5

| Hyperparameter | Value |
|---|---|
| Batch size | 32 |
| Segment length | 64 |
| Optimizer | Adam |
| Learning rate | $2 \times 10^{-5} \to 1 \times 10^{-6}$ |
| Learning rate scheduler | Exponential decay |

Table 3: Common training hyperparameters for MetaWorld and D4RL benchmarks.

| Hyperparameter | Task | Value |
|---|---|---|
| Regularization weight $\mu$ | Bin Picking | 1.0 |
| | Button Press | 1.0 |
| | Door Open | 1.0 |
| | Drawer Open | 1.0 |
| | Plate Slide | 0.1 |
| | Sweep into | 1.0 |
| | Walker2d | 2.0 |
| | Hopper | 2.0 |
| | HalfCheetah | 2.0 |
| | pen | 2.0 |
| | kitchen | 2.0 |
| Bias $b$ | Bin Picking | 0 |
| | Button Press | 0 |
| | Door Open | 0 |
| | Drawer Open | 0 |
| | Plate Slide | 0 |
| | Sweep into | 0 |
| | Walker2d | 0 |
| | Hopper | 0 |
| | HalfCheetah | 0 |
| | pen | 0 |
| | kitchen | 0 |
| Temperature $\rho$ | Bin Picking | 500 |
| | Button Press | 500 |
| | Door Open | 500 |
| | Drawer Open | 500 |
| | Plate Slide | 500 |
| | Sweep into | 500 |
| | Walker2d | 500 |
| | Hopper | 500 |
| | HalfCheetah | 500 |
| | pen | 500 |
| | kitchen | 500 |

Table 4: Specific hyperparameters of FKPD for each task.

## 4.3 Computation Resource

We carry out our experiment on NAVIDIA RTX 4090 GPU server. For each MetaWorld or D4RL task, We trained our policy for 3 hours, completing 400k steps of iteration.


\end{document}